\documentclass{llncs}

\usepackage[utf8]{inputenc}
\usepackage[T1]{fontenc} 
\usepackage{hyperref}
\usepackage{bm}
\usepackage{color}
\usepackage{amsmath,amsfonts}
\usepackage{graphicx}
\usepackage{tikz}
\usepackage{xspace}
\usetikzlibrary{shapes,arrows}

\tikzstyle{decision} = [diamond, draw,
    text width=4.5em, text badly centered, node distance=3cm, inner sep=0pt]
\tikzstyle{block} = [rectangle, draw,
    text width=6em, text centered, rounded corners, minimum height=4em]
\tikzstyle{line} = [draw, -latex']
\tikzstyle{endnode} = [draw, ellipse, node distance=3cm,
    minimum height=2em]

\newcommand\myparagraph[1]{\vspace{0.1cm} \parindent0pt \textbf{#1}\quad}

\title{Neither Quick Nor Proper -- Evaluation of QuickProp for Learning Deep Neural Networks}
\author{Clemens-Alexander Brust \and Sven Sickert \and Marcel Simon \and Erik Rodner \and Joachim Denzler}
\institute{Computer Vision Group, Friedrich Schiller University Jena, Germany\\\url{http://www.inf-cv.uni-jena.de}}

\begin{document}

\maketitle

\begin{abstract}
  Neural networks and especially convolutional neural networks are of great interest in current computer vision research.
  However, many techniques, extensions, and modifications have been published in the past, which are not yet used by current approaches.
  In this paper, we study the application of a method called QuickProp for training of deep neural networks.
  In particular, we apply QuickProp during learning and testing of fully convolutional networks for the task of semantic segmentation.
  We compare QuickProp empirically with gradient descent, which is the current standard method.
  Experiments suggest that QuickProp can not compete with standard gradient descent techniques 
  for complex computer vision tasks like semantic segmentation.
\end{abstract}

\section{Introduction}
\label{sec:intro}
Convolutional neural networks (CNNs) \cite{LeCun89:BAH} achieve state-of-the-art performance in many areas of computer vision including image classification, object detection, and segmentation \cite{He15:DRL,Krizhevsky12:ICD,Long15:FCN,Simon15:NAC}.
However, the training time of large networks is a major bottleneck when evaluating different architectures and new types of layers.
In this work, we study the QuickProp algorithm \cite{Fahlman88:ESL} for decreasing training time in the domain of semantic segmentation and potentially reaching a better optimum during training. 

QuickProp \cite{Fahlman88:ESL} is a second-order optimization algorithm that uses a simple approximation of the Hessian's diagonal to accelerate optimization and therefore
belongs to the class of Quasi-Newton algorithms.
So far, it has been evaluated and studied only for standard neural network training~\cite{Magoulas99:ICB,Vrahatis99:CAQ,Vrahatis00:GCM}.
However, current
neural network architectures, like CNNs, are characterized by a significantly larger number of parameters. Furthermore, the increased number of layers leads to a higher
numerical error when computing the gradients with back-propagation~\cite{Hochreiter01:GFR}.
For an evaluation of its performance, we test the algorithm in synthetic and real-world experiments and compare its behavior to the traditional and widely used optimization approach gradient descent (GD).

In this work, we focus on the task of semantic segmentation.
Each pixel of an image is classified into known classes and thereby the image is segmented into meaningful regions.
In contrast to unsupervised segmentation, this not only returns possible object boundaries as regions but also semantic labels for each of those regions.
For this task, we make use of the convolutional network architectures as proposed in~\cite{Brust15:CPN} and use them throughout all our experiments.

The remainder of the paper is structured as follows. 
In Section~\ref{sec:related}, we give a brief review of related work and optimization techniques for the task of network training.
In Section~\ref{sec:quickprop}, we elaborate on the QuickProp algorithm which was originally introduced in~\cite{Fahlman88:ESL}.
An evaluation using synthetic and real-world experiments is given in Section~\ref{sec:experiments}.
A discussion and summary in Section~\ref{sec:conclusion} concludes the paper.

\section{Related work}
\label{sec:related}

    The most related approach to QuickProp is Newton optimization, where the Hessian of the objective function is used for a second-order approximation.
    However, computing the Hessian exactly as done in the Newton method is infeasible for neural networks with more than a few thousand parameters. For example,
    current state-of-the-art architectures in the computer vision area usually parameters in the order of millions~\cite{Krizhevsky12:ICD}.
    Therefore, Quasi-Newton methods are required approximating the Hessian in a computationally efficient manner. In fact, QuickProp uses a simple approximation that
    comes with nearly no additional computational cost. Other alternative Quasi-Newton methods are reviewed in \cite{Martens12:TDR}.

    To speed up QuickProp even further, the work of \cite{Cheung12:IQA} magnifies the gradient of sigmoid activation layers inspired by the technique proposed by \cite{Ng04:MGF}.
    This method has no effect on the learning with state-of-the-art CNN architectures, since rectified linear units are used, whose gradients are either zero or one and 
    can therefore not be exponentially magnified.
    Another modification is given by \cite{Vrahatis99:CAQ,Vrahatis00:GCM}, where absolute differences of the gradients and former weight changes are used and multiplied with a tuning parameter $\eta$.
    The authors also show the convergence of QuickProp for continuously differentiable objectives satisfying several additional conditions also for the gradients.
    A comparision of several optimization algorithms including QuickProp is given by \cite{Magoulas99:ICB}, but limited to standard neural networks with a relatively small number of parameters. 

    A very common optimization strategy used for training deep neural networks is gradient descent with momentum as studied in \cite{Sutskever13:IIM}. Similar to QuickProp, momentum incorporates the change
    of weights of the previous iteration. However, this is done in an additive manner, whereas QuickProp performs a multiplicative update and also handles each component of the 
    weight vector independently.
    Other methods including AdaDelta~\cite{zeiler12adadelta}, AdaGrad~\cite{duchi11adagrad} and Adam~\cite{kingma14adam} try to improve over gradient descent by estimating a optimal learning rate in each update step.
    In contrast, QuickProp assumes the output to be a quadratic function of each variable and calculates the location of the optimum directly.

\section{QuickProp optimization algorithm}
\label{sec:quickprop}
Quick Propagation or QuickProp~\cite{Fahlman88:ESL} is an iterative second-order optimization algorithm. 
The algorithm approximates in each step the objective function with a quadratic function for each variable independently from the others.
In the case of neural networks, the input variables of the objective functions are the parameters or weights of the network. 

\myparagraph{QuickProp optimization step}
Let $E(\bm{w})$ be the error (or loss) function of a neural network with respect to the parameters $\bm{w}=(w_1,w_2,\dots,w_K)$.
The goal of training is to minimize $E(\bm{w})$ for the training set.

In each iteration, QuickProp determines the new weight $\bm{w}^{(t+1)}$ given the current weight $\bm{w}^{(t)}$ as well as the slope $\vec{g}=\frac{\partial E}{\partial \bm{w}}$ of the previous and current step.
Gradient descent updates the weights by calculating $\bm{w}^{(t+1)} = \bm{w}^{(t)} + \Delta \bm{w}^{(t-1)} = \bm{w}^{(t)}+\gamma \cdot \vec{g}^{(t)}$ with $\gamma \in \mathbb{R}$ being the
learning rate or step length and $\vec{g}^{(t)} = \nabla_{\bm{w}} E(\bm{w}^{(t)})$ being the gradient of $E$ evaluated at $\bm{w}^{(t)}$.
In QuickProp, each weight $w_i$ is optimized independently and given a selected component $i$, all weights except $w^{(t)}_i$ are assumed to be fixed.
We therefore assume $\bm{w}$ to be a scalar $w$ in the following. To derive an update rule for $w^{(t+1)}$, $E$ is approximated by a parabola using a second-order Taylor approximation assuming
that $E$ is twice differentiable:
\begin{align}
E(w^{(t+1)}) &\approx E(w^{(t)}) + g^{(t)} \cdot \Delta w^{(t)} + \frac{1}{2} E''(w^{(t)}) \left(\Delta w^{(t)}\right)^2
\end{align}
Whereas this is a standard approximation used by all (Quasi-)Newton methods, QuickProp now approximates the second derivative $E''$ by a finite difference using $w^{(t)}$ and $w^{(t-1)}$:
\begin{align}
E''(w^{(t)}) &\approx \frac{g^{(t)} - g^{(t-1)}}{\Delta w^{(t-1)}}
\end{align}
In contrast to the Hessian-free optimization technique presented in \cite{Martens12:TDR}, this does not require an additional pass through the network for
forward differentiation. However, it savings in computation time come with greater numerical instabilities that need to be tackled and are further discussed in the next section.

By combining both equations and computing the stationary point of the approximation of $E$ with respect to $\Delta w^{(t)}$, we get:
\begin{equation}
\label{eq:qp}
\Delta w^{(t)} = \frac{g^{(t)}}{g^{(t-1)}-g^{(t)}} \cdot \Delta w^{(t-1)}
\end{equation}
which is the update step used by QuickProp.
The local parabola approximation $p$ of QuickProp around $w^{(t)}$ can be therefore expressed by:
\begin{align}
p(z) &= a \cdot (z-w^{(t)})^2 + b \cdot (z-w^{(t)}) + c\mbox{, where}\\
a &= \frac{1}{2} \frac{g^{(t)} - g^{(t-1)}}{\Delta w^{(t-1)}}\\
b &= g^{(t)}\\
c &= E(w^{(t)})
\end{align}

\myparagraph{Special cases}
Instead of adding $\Delta w^{(t)}$ immediately to the weights, the QuickProp algorithm handles the following three cases separately:
\begin{enumerate}
  \item{\textbf{The absolute value of the current slope is smaller than the previous one, but has the same sign}:
  In this case, the new weight is obtained by adding the delta, \textit{i.e.} $w^{(t+1)}=w^{(t)}+\Delta w^{(t)}$.
  The step $\Delta w^{(t)}$ depends on the difference between the current and previous slope as described in eq.~\ref{eq:qp}.}
  
  \item{\textbf{The signs of the slopes differ:}
  The minimum lies between the current and the previous weight $w^{(t-1)}_i$ and $w^{(t)}_i$.
  The update rule from eq.~\ref{eq:qp} results in a backwards step because of the different signs.}
  
  \item{\textbf{The absolute value of the current slope is larger or equal to the previous slope}:
  This would result in an infinite step or a step in the wrong direction.
  The step size needs to be limited.
  The author introduces the hyperparameter $\mu$, called the ``maximum growth factor''.
  A weight step can never be larger than $\mu$ times the previous step.
  Steps that are too large or infinite or in the wrong direction are replaced by $\mu$ times the previous step.
}
\end{enumerate}
To initialize the algorithm, a traditional gradient descent step is taken.
Gradient descent is also used for weights that did not change in an iteration, which would result in $\Delta w^{(t)}$ being $0$ for all following time steps.

\myparagraph{Implementation details}
	Our implementation follows exactly the work of Fahlman~et~al.~\cite{Fahlman88:ESL}.
	To ignite the optimization process or when the gradient becomes too small,
	we use a single fixed learning rate gradient descent step. We set this threshold to $1e-15$.
	The implementation also uses the maximum growth factor and adds a gradient step if the current and the previous
	gradient have the same signs.

\begin{figure}[t]
  \centering
  \includegraphics[width=.9\textwidth]{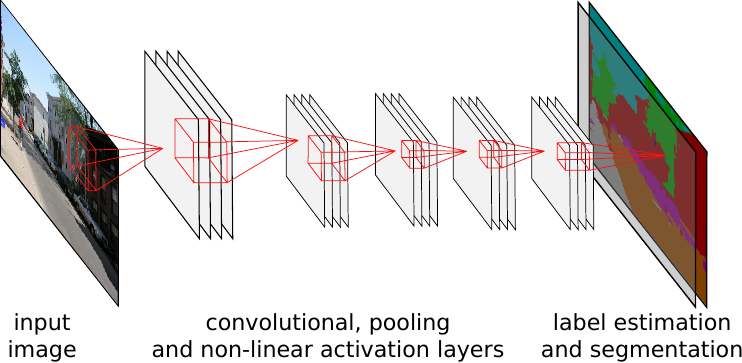}
  \caption{Architecture of convolutional neural networks used for semantic segmentation.}
  \label{fig:cnnfully}
\end{figure}

\section{Experiments}
\label{sec:experiments}

In order to evaluate the QuickProp algorithm in a modern context we use two datasets of different complexities for the task of semantic segmentation.
In this section, we investigate how QuickProp compares to gradient descent which is the standard technique for training neural networks.
After an introduction of the datasets and a description of the experimental setup, we analyze the influence of the network complexity.%

\subsection{Deep neural networks for semantic segmentation}

Our experiments focus on the application of deep neural networks for semantic segmentation using so-called fully-convolutional neural networks~\cite{Long15:FCN}.
The goal of semantic segmentation is to classify each pixel of a given image into one of $K$ semantic categories. 
\figurename~\ref{fig:cnnfully} shows the basic principle of the architecture of a CNN used for this task.
The input of the network is a given image and the output is a score for each class and pixel.
The index of the maximum score then corresponds to the semantic class assigned to the pixel.

The network is comprised of different layers, mainly (a) convolutional layers, (b) pooling layers, (c) up-sampling layers, and (d) non-linear activation layers.
Since we do not focus on the application of semantic segmentation in our paper but rather on efficient optimization techniques for learning, we
refer the interested reader to \cite{Brust15:CPN,Long15:FCN} for more details.

\subsection{Experimental setup}

In order to evaluate the performance of the optimization techniques for semantic segmentation tasks, we report typical quantitative measures in this field of research.
With respect to the semantic labeling of pixels we follow \cite{Brust15:CPN} in measuring the performance using the overall accuracy and mean class-wise accuracy.
Furthermore, we analyze the different optimization methods using the output of the quadratic loss function.
More specifically, the reported \textit{loss} in our experiments is the per pixel error averaged over all pixels of all examples and weighting updates.
Additionally, all results shown in this section are averaged over several runs allowing for a more robust evaluation of the used techniques.

\subsection{Datasets}

\begin{figure}[t]
  \centering
  \begin{minipage}{0.13\textwidth}
    \includegraphics[width=1.0\textwidth]{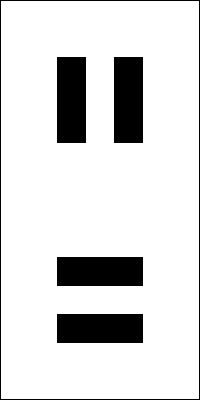}
  \end{minipage}
  \begin{minipage}{0.13\textwidth}
    \includegraphics[width=1.0\textwidth]{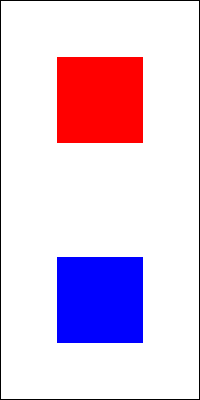}
  \end{minipage}
  \hfill
  \begin{minipage}{0.345\textwidth}
    \includegraphics[width=1.0\textwidth]{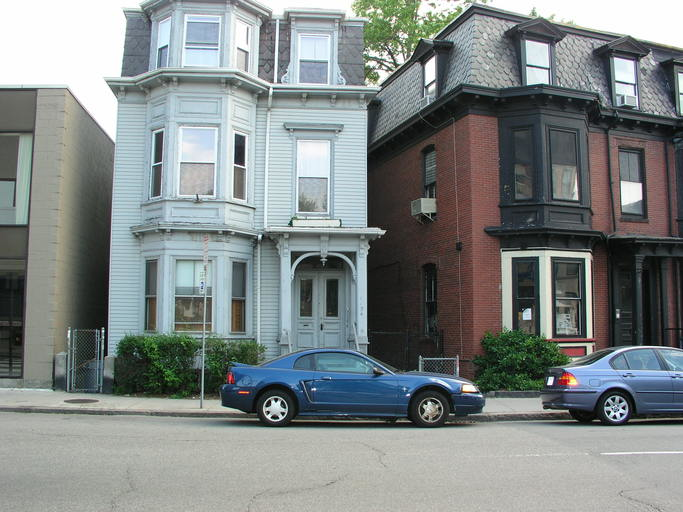}
  \end{minipage}
  \begin{minipage}{0.345\textwidth}
    \includegraphics[width=1.0\textwidth]{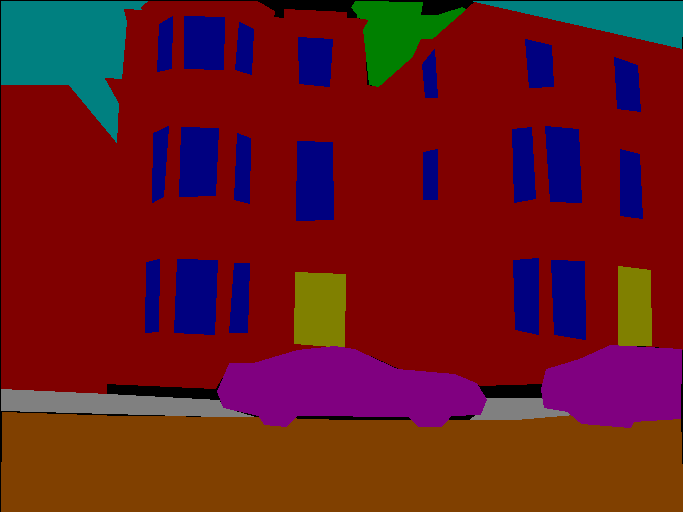}
  \end{minipage}
  \caption{Datasets used in our experiments: A simple three class scenario in the T.O.Y. dataset (left) and a multi-class setting of urban scenes in the LabelMeFacade dataset \cite{Froehlich12:SSM} (right). Depicted are both input image and the corresponding ground truth labeling as color coding.}
  \label{fig:datasets}
\end{figure}

For our experimental evaluation we used the following datasets which have differentrequirements on the type and complexity of the network architecture.

\myparagraph{T.O.Y. dataset}
    In order to analyze the performance for varying settings we created a small artificial dataset serving as a working example.
    This so-called T.O.Y. dataset consists of only a single image with two foreground classes and one background class.
    It is designed to be easy to solve even for a tiny network.
    The training image and its corresponding labeling is depicted in Fig. \ref{fig:datasets}.
    The T.O.Y. dataset allows for the verification whether QuickProp yields reasonable solutions because the classes are differentiable by basic filter masks.
    A second advantage of this dataset is the small size.
    There is a trend to publish the results of just a single random initialization of a CNN, especially in competitive settings.
    However, analyzing the distribution over many repetitions provides results with statistical significance.

\myparagraph{LabelMeFacade}
    The second dataset we use in our experimental evaluation is LabelMeFacade \cite{Froehlich12:SSM} which consists of 945 images taken from urban scenes.
    It contains ground truth labels for eight classes: \textit{building}, \textit{road}, \textit{pavement}, \textit{sky}, \textit{vegetation}, \textit{window} and \textit{door}.
    Additionally, there exists a background class which can be used to exclude undefined areas from the evaluation.
    Fig. \ref{fig:datasets} shows an example containing all labeled classes with their corresponding color coding.

    In our evaluation we follow \cite{Froehlich12:SSM} and split the dataset into 845 images for testing and 100 images for training.
    For model selection and tuning of hyperparameters, we split the 100 training images again into 50 images each for training and 50 images for validation.

\subsection{Comparison of QuickProp and gradient descent}

\myparagraph{T.O.Y. dataset}
    We train a network 200 times from scratch with different random initializations.
    The training is stopped after 30 ``epochs'' of 10.000 iterations each and one image per iteration. 
    The CNN architecture is a simple 2-layer CNN. 
    One convolutional layer with two filter masks are followed by output neurons and hence corresponds to a simple template matching approach.
    The input to the network is exactly as large as the filters in the first layer. 
    A quadratic loss is used in order to match the assumptions of QuickProp as close as possible. 

\begin{figure}[t]
      \centering
      \begin{minipage}[c]{0.49\textwidth}
	\centering
	\includegraphics[width=1.\textwidth]{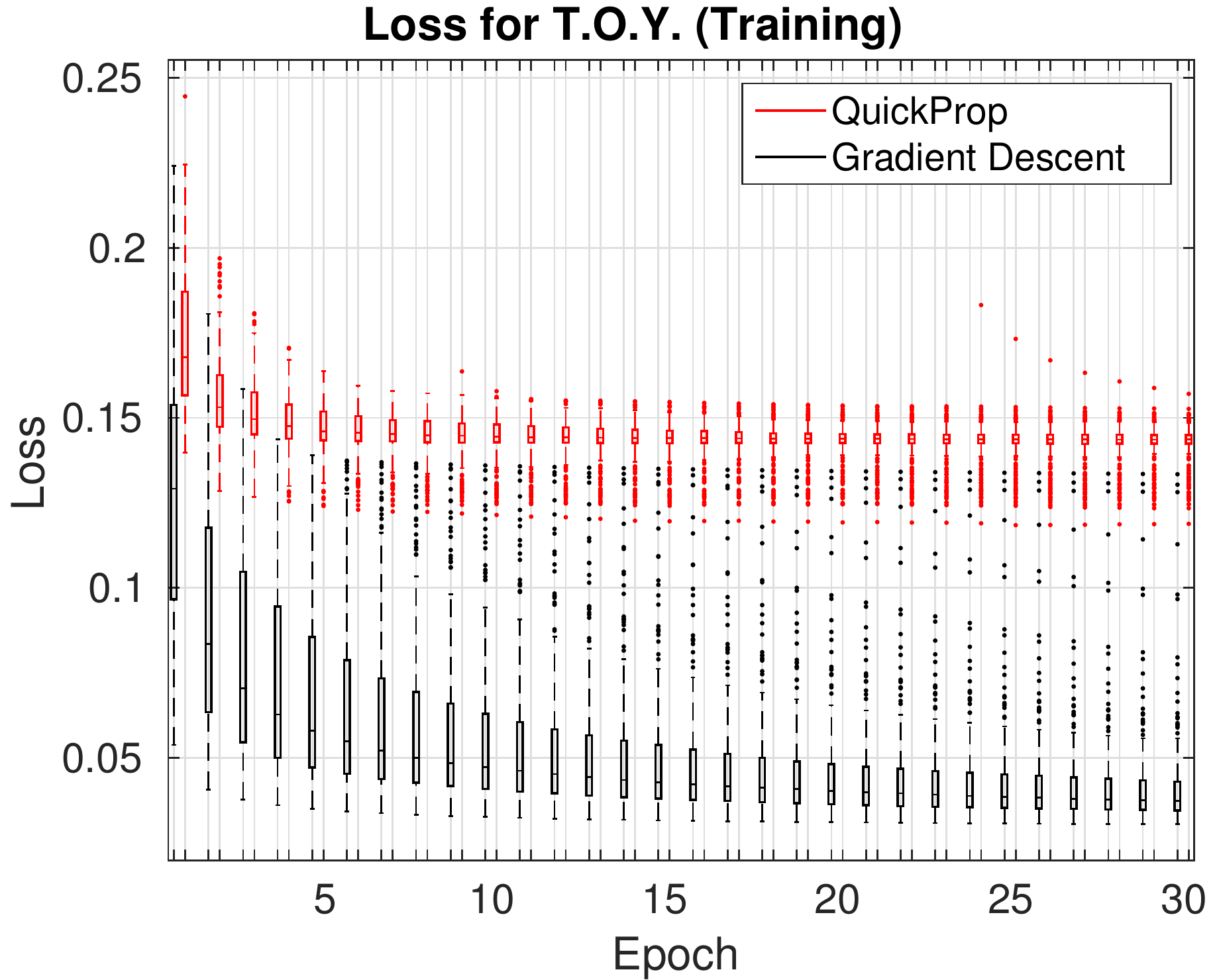}
      \end{minipage}
      \hfill
      \begin{minipage}[c]{0.49\textwidth}
	\centering
	\includegraphics[width=1.\textwidth]{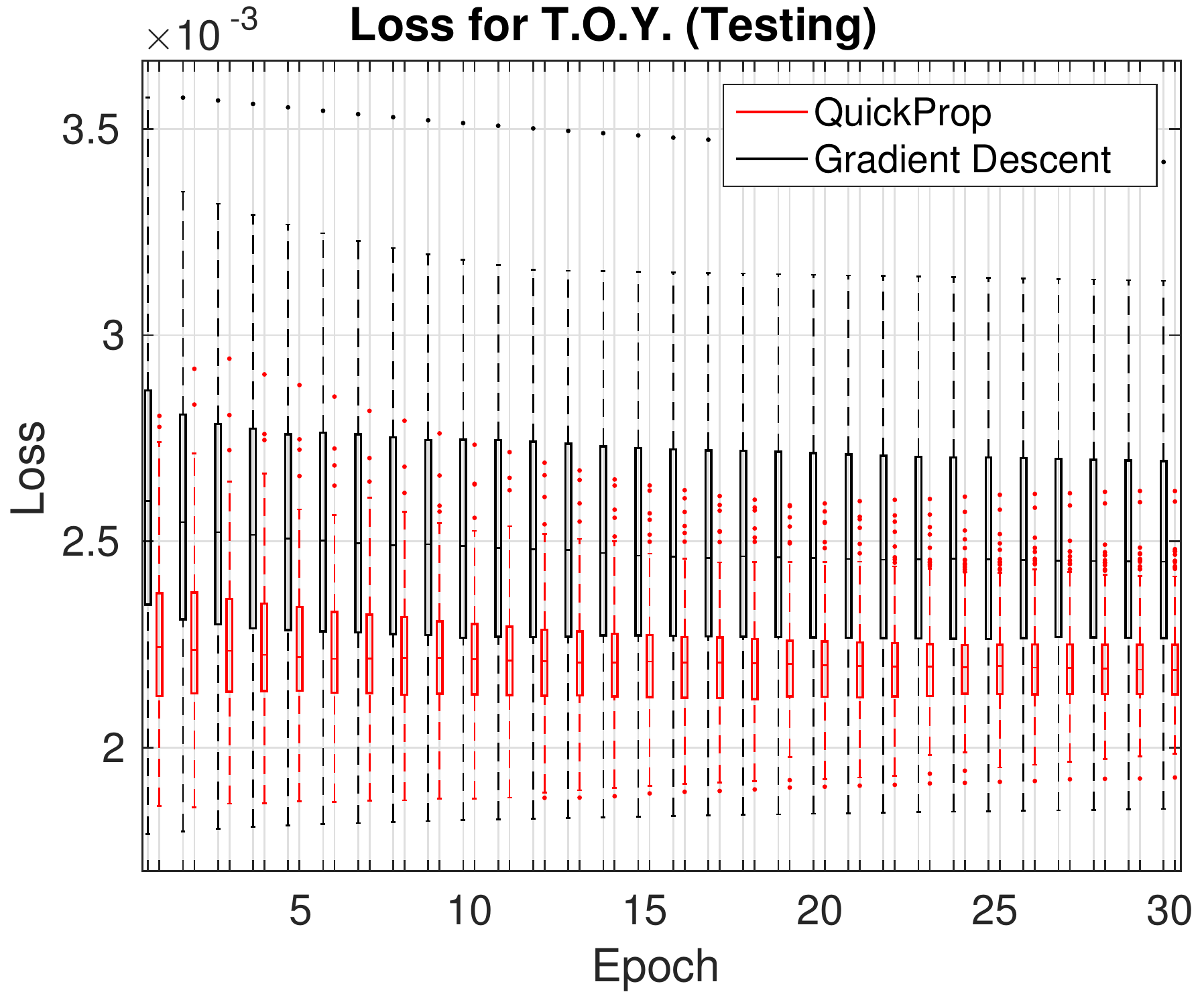}
      \end{minipage}
      \caption{Quantitative analysis using the T.O.Y. dataset: during training the loss using QuickProp is substantially worse than with gradient descent (left). However, during testing the relationship is inverted (right).}
      \label{fig:toy_epochs}
    \end{figure}

    In Fig.~\ref{fig:toy_epochs} we show how the loss is decreasing during training and subsequent testing of the networks after each epoch.
    It can be seen that in training QuickProp is outperformed by gradient descent.
    The results during testing, however, are inverted with respect to the reported loss.
    This behavior is interesting, since it suggests a better generalization ability of the network which was trained using QuickProp.
    However, the visual variety of the T.O.Y. dataset is very small and hence a general statement is not possible.

    The overall accuracy of the learned networks during testing was comparable.
    It saturated at 78\% for both the gradient descent and the QuickProp approach after a few epochs.
    In contrast, the mean class-wise accuracy for QuickProp was significantly worse compared to the standard learning approach.
    While the gradient descent technique saturated at 62\% after five epochs, QuickProp never rised above 41\%.

\myparagraph{LabelMeFacade dataset}
    The architecture used for this set of experiments consists of three convolutional layers and two fully connected layers.
    Since we are using a fully convolutional setting the fully connected layers are also convoltional layers of kernel size $1\times1$ \cite{Long15:FCN}.
    Each convolutional layer is followed by subsequent non-linearity layer.
    While the last non-linear layer is \texttt{Sigmoid}, the remaining activation functions are \texttt{Tanh}.
    We trained CNNs for ten epochs for each optimization method.
    The training process is repeated ten times.

\begin{figure}[t]
      \centering
      \begin{minipage}[c]{0.31\textwidth}
	\centering
	\includegraphics[width=1.\textwidth]{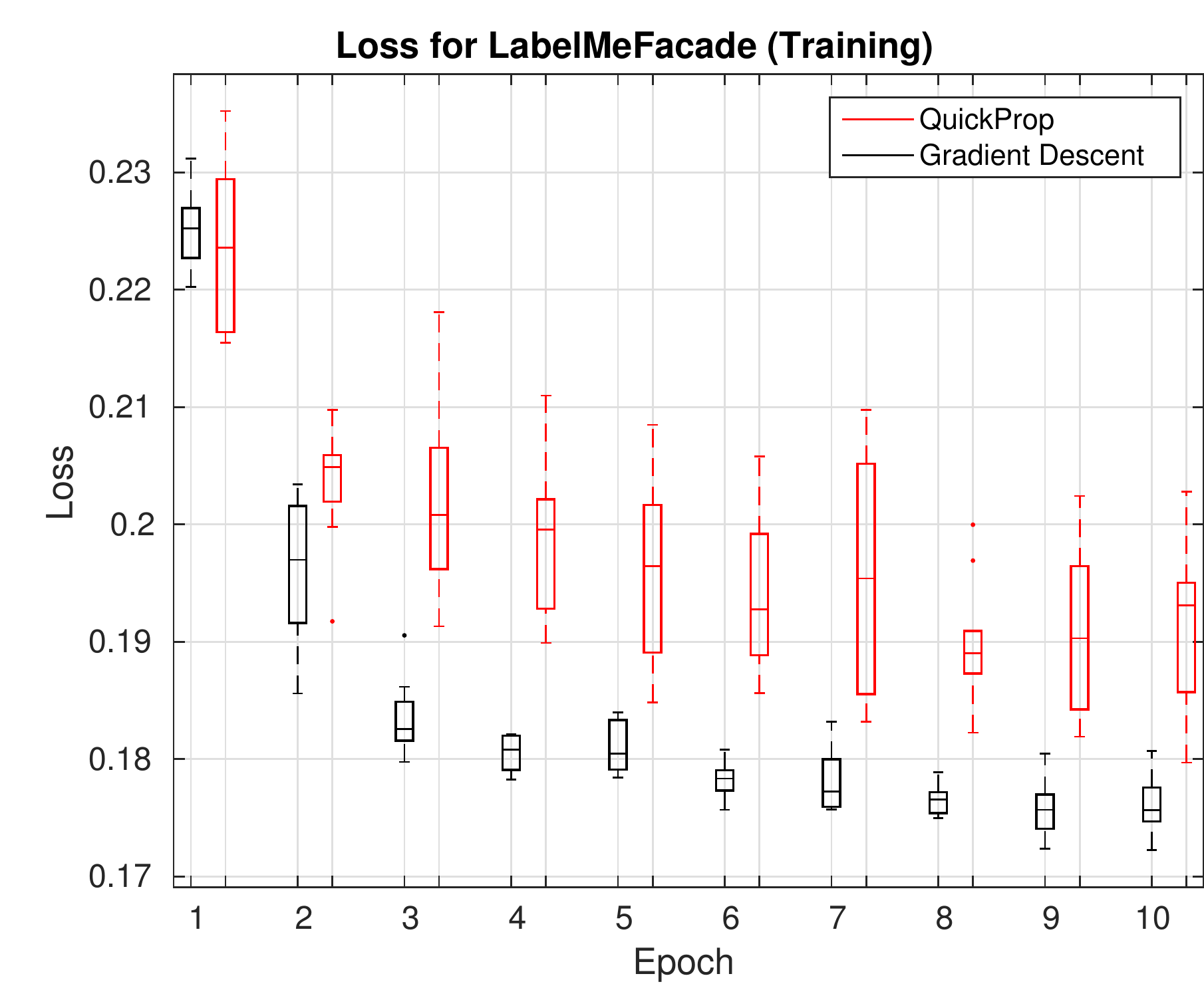}
      \end{minipage}
      \hfill
      \begin{minipage}[c]{0.31\textwidth}
	\centering
	\includegraphics[width=1.\textwidth]{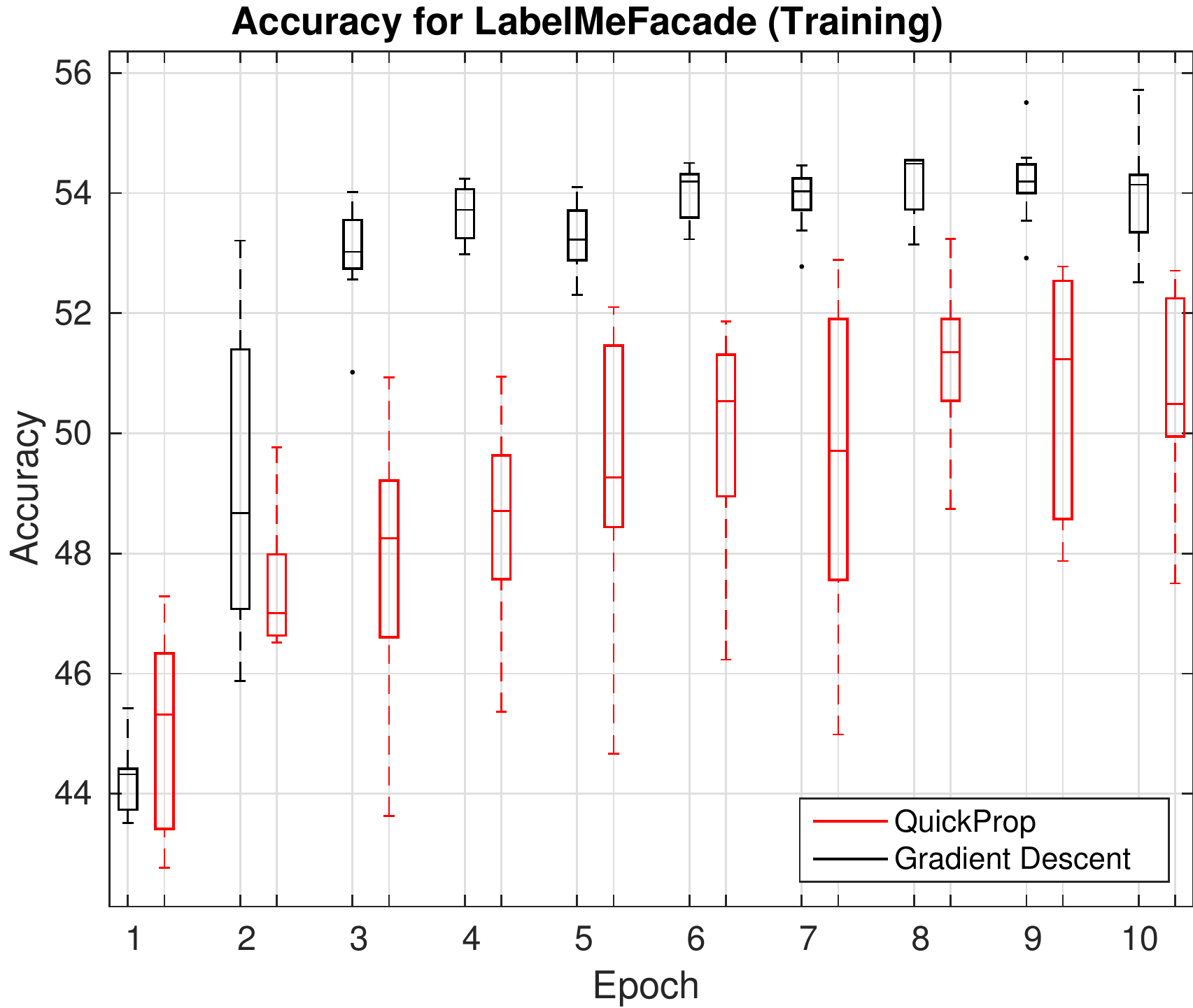}
      \end{minipage}
      \hfill
      \begin{minipage}[c]{0.31\textwidth}
	\centering
	\includegraphics[width=1.\textwidth]{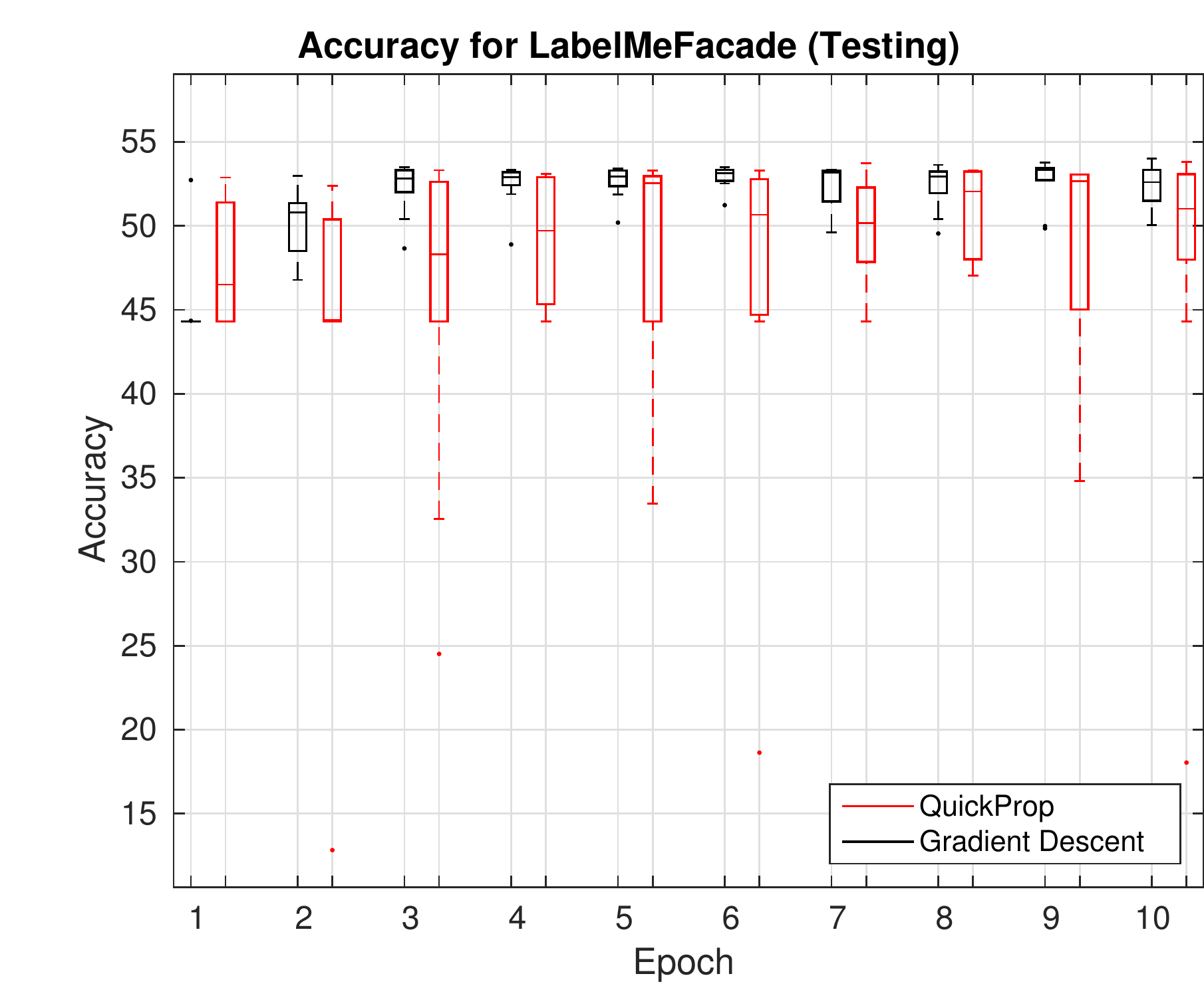}
      \end{minipage}
      \caption{Quantitative analysis using LabelMeFacade: During training and testing gradient descent achieves better results than QuickProp.}
      \label{fig:lmf_epochs}
    \end{figure}

    Fig.~\ref{fig:lmf_epochs} shows the loss over time during training as well as the accuracy during training and testing.
    As can be seen, QuickProp performs worse than gradient descent in this scenario.
    For each training epoch the error rates of Quickprop are higher and the resulting accuracy is up to five percent points below gradient descent.
    The reported accuracy for QuickProp during testing is also worse.
    Additionally, an increase in variance of the results can be observed.

    In comparison to earlier experiments, the results for testing do not support the findings of the T.O.Y. experiments with respect to the generalization ability.
    However, it is worth noting that the LabelMeFacade dataset iFs in general more complex than the T.O.Y. dataset.
    While the latter is a simple binary classification task the former features a multi-class setting with a high variety of visual representations for the classes.
    As a consequence, the network in use was of higher complexity than the one for the T.O.Y. dataset.
    In order to investigate whether the architecture of the CNN plays a defining role in the performance of QuickProp we ran a second series of experiments on LabelMeFacade.

\subsection{Network complexity analysis}
    The models studied so far range from a very simple network of only 109 parameters for the T.O.Y. dataset up to networks featuring more than 100.000 weights for LabelMeFacade.
    In this set of experiments we evaluate if the model complexity relates to the performance of the different training schemes.
    We propose two scalable network architectures based on the initial model we used for the LabelMeFacade experiments.
    The first analysis features a variable count of filters in a specific layer.
    After that, the network is modified in a way to enable a variable amount of layers.

\myparagraph{Scaling the amount of filters}
    First, we vary the number of filters in the second convolutional and the first fully connected layer.
    Both layers are linearly dependent on a variable $k$, using $k$ filters for convolution and $12\cdot k$ neurons in the fully connected layer.
    Accordingly, the total amount of parameters $|\mathbf{w}|$ in the networks can then be expressed in terms of $k$ as $|\mathbf{w}|=2473+4497\cdot k$.
    In order to evaluate the influence of differently sized networks we train and test CNNs on the LabelMeFacade dataset for $k\in\{2,7,12,\dots,22\}$.

\begin{figure}[t]
      \centering
      \begin{minipage}{0.49\textwidth}
        \centering
        \includegraphics[width=1.\textwidth]{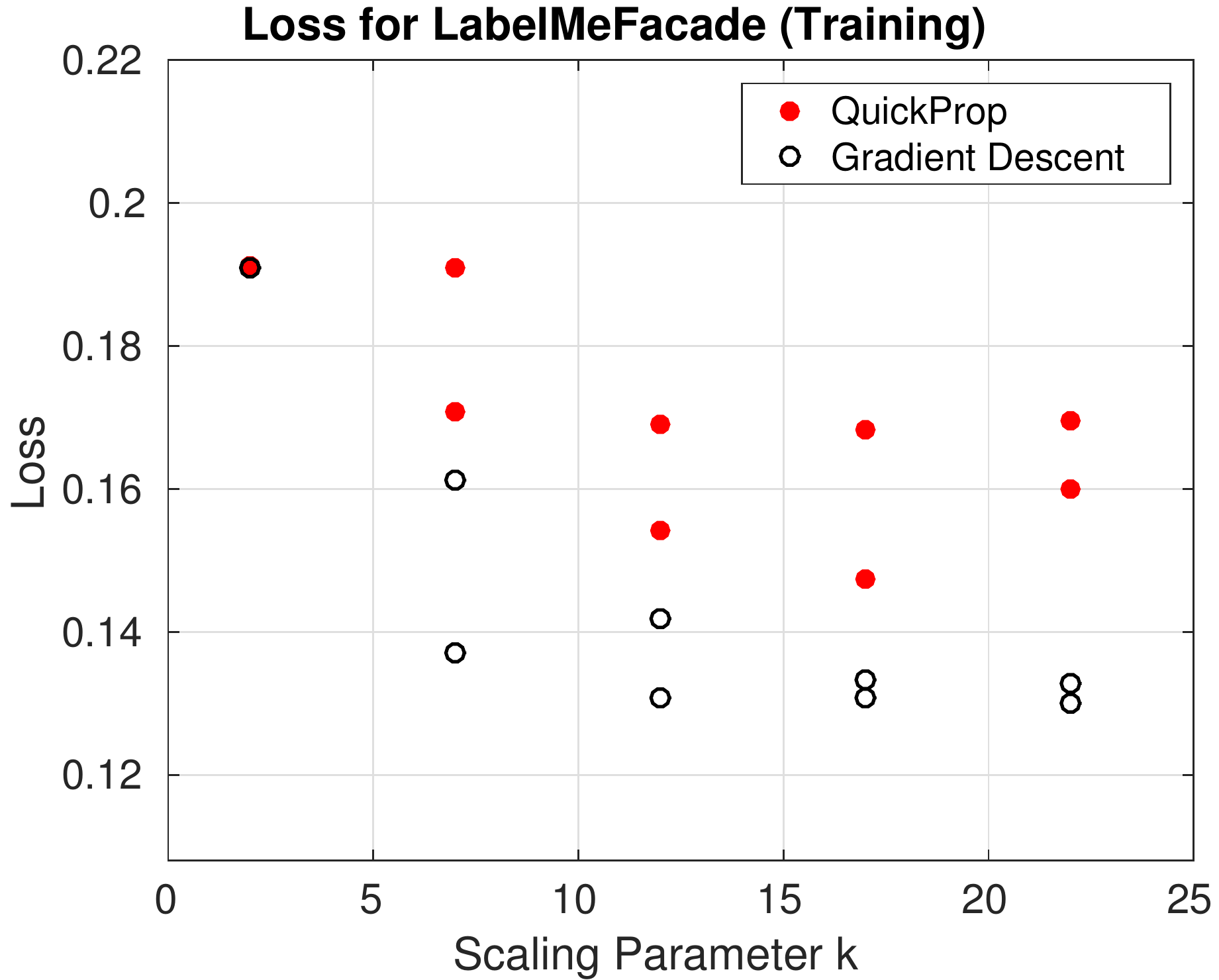}
      \end{minipage}
      \hfill
      \begin{minipage}{0.49\textwidth}
        \centering
        \includegraphics[width=1.\textwidth]{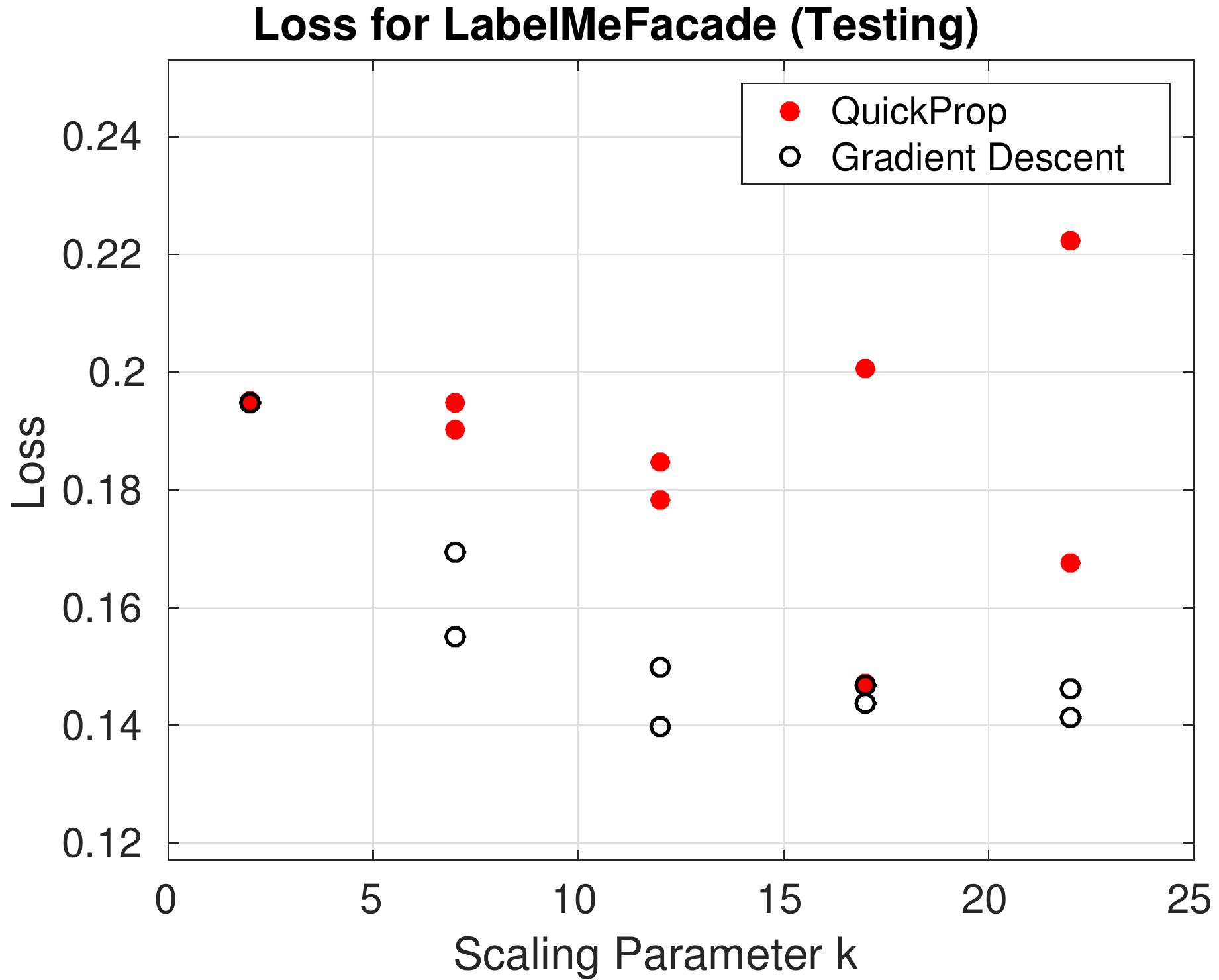}
      \end{minipage}
      \caption{Scaling the amount of filters per layer to increase network complexity: }
      \label{fig:horiz_loss_comp}
    \end{figure}

    In Fig.~\ref{fig:horiz_loss_comp} we report the loss reached for networks of increasing size.
    As before, we compare networks trained using gradient descent and QuickProp.
    For each $k$ the CNNs were learned for 100 epochs.
    Since the training took a very long time, we were only able to repeat the training once for each setup.
    
    As can be seen from the results in Fig.~\ref{fig:horiz_loss_comp}, the loss during training for QuickProp is higher than using GD for almost all cases.
    Interestingly, the loss for the case $k=2$ is the same for all runs.
    However, a general increase of the gap between both methods is visible for more complex networks.
    This supports the findings that QuickProp performs better for smaller networks in this application.

\begin{figure}[t]
      \centering
      \begin{minipage}{0.49\textwidth}
        \centering
        \includegraphics[width=1.\textwidth]{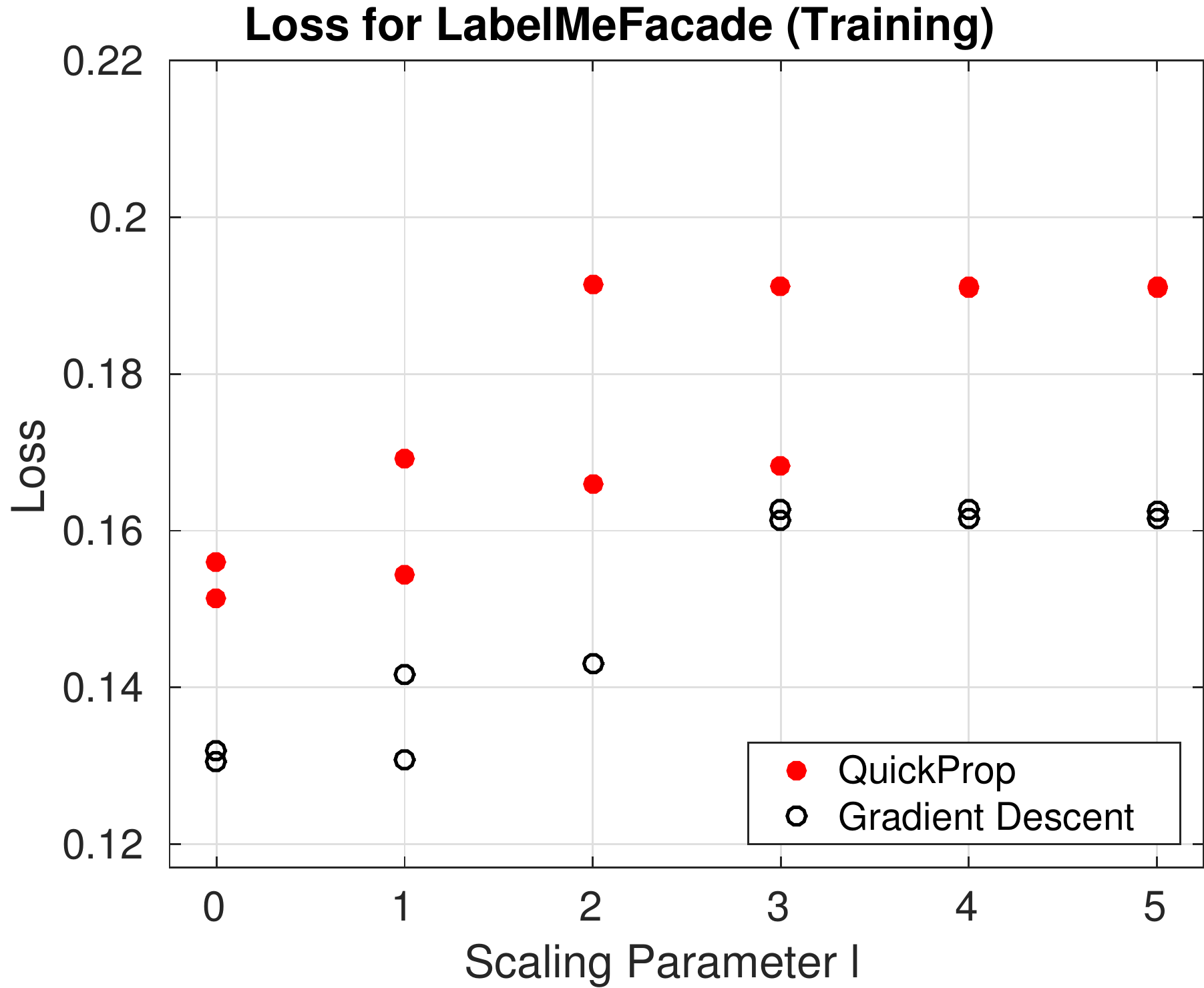}
      \end{minipage}
      \hfill
      \begin{minipage}{0.49\textwidth}
        \centering
        \includegraphics[width=1.\textwidth]{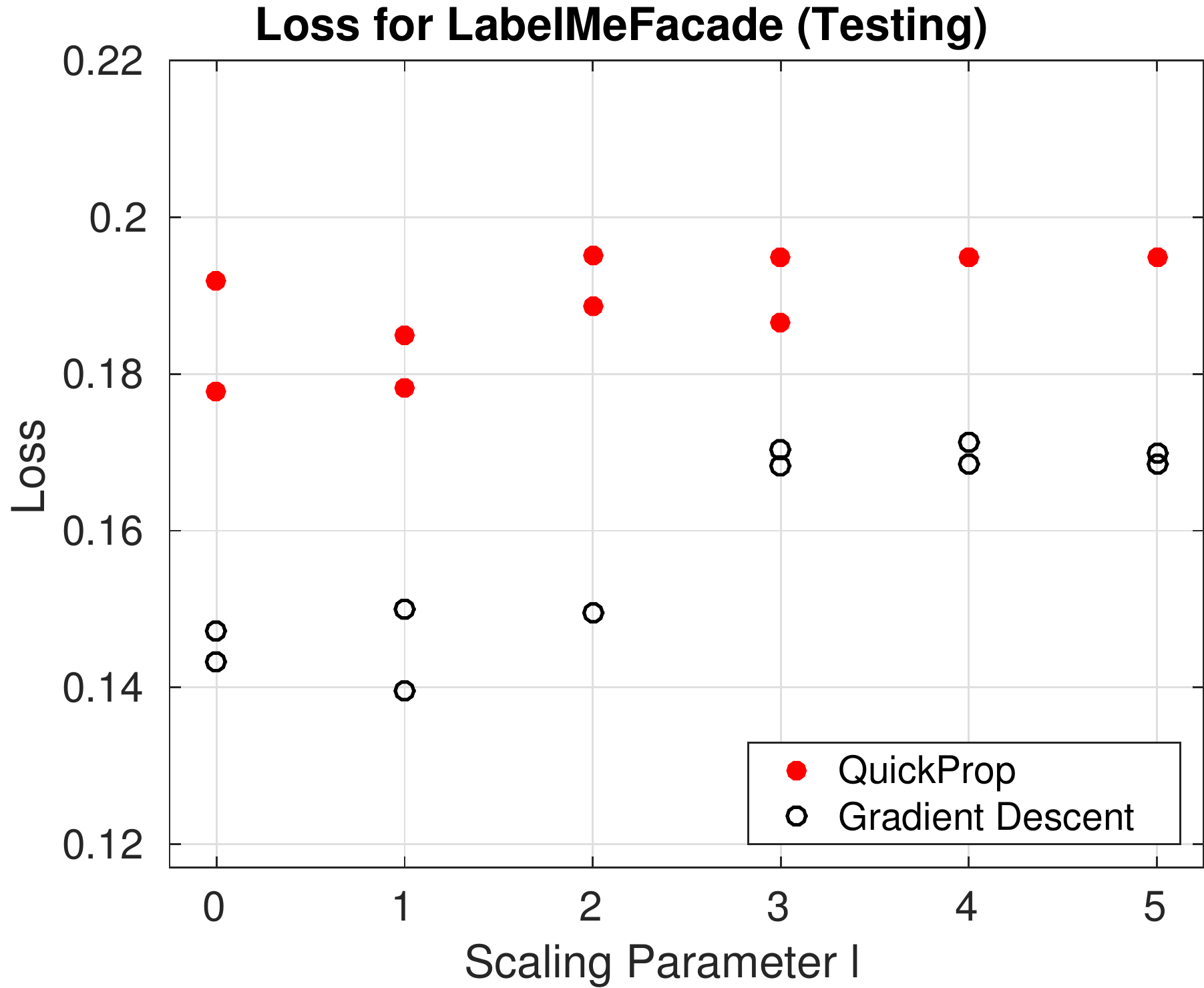}
      \end{minipage}
      \caption{Scaling the amount of layers to increase network complexity: }
      \label{fig:vert_loss_comp}
    \end{figure}

\myparagraph{Scaling the amount of layers}
    Another way of increasing the complexity of CNNs is to add more layers to its architecture.
    In order to achieve a linear growth in weights, we suggest to repeat the first fully connected layer with 192 kernels $l$ times.
    As this layer is in fact a convolutional layer of size $1\times1$, the total amount of parameters in the networks computes as $|\mathbf{w}| = 56437 + l$, with $l\in\{0,1,\dots,5\}$ being a linear scaling parameter.
    Experiments were conducted for a step-wise increased $l$ and we compared the same optimization methods as before.
    We repeated each experimental setup once and report both runs in Fig.~\ref{fig:vert_loss_comp}.
    
    Similar to our earlier experiments, QuickProp performs worse when compared to the standard technique gradient descent.
    However, in comparison to the scaling of the amount of filters, the gap between the performance of both methods is not increasing.
    On the contrary, results during testing show a decreasing gap.
    
    However, it is worth noting that a simple repetition of the first fully connected layer seems to be unreasonable.
    For an increasing $l$ the loss is getting worse and for $l>2$ stays constant in our series of experiments.
    Also note, that without losing performance the first fully connected layer can be skipped altogether ($l=0$).
    This might be due to the fully convolutional character of our network.

\section{Conclusions}
\label{sec:conclusion}

In this technical report, we evaluated QuickProp as an iterative second-order optimization algorithm for the training of convolutional neural networks.
We conducted experiments for the task of semantic segmentation using a fully convolutional network configuration.

For an inital evaluation, a simple toy example as well was used allowing a high amount of repetitions of the experiments in order to retrieve results of statistical significance.
While the QuickProp algorithm showed a higher loss during training, the performance during testing was superior to a network trained using gradient descent.

However, for a \textit{real-world} scenario using a dataset consisting of urban street sceens, the standard approach of gradient descent still performs significantly better.
In order to investigate a possible relationship between the size and complexity of networks to be trained and the performance of optimization techniques, we ran scaling experiments of two different types.
The series of experiments for a scalable amount of filter kernels in selected layers showed a degrading performance of QuickProp for larger networks when compared to gradient descent.

In general, improved second-order optimization approaches, like QuickProp, can help in speeding up the training process and finding a better optimum.
However, these cases seem to be rare and do not include practical scenarios featuring convolutional neural networks.
Since the models of recently propsed networks for computer vision task become larger, we can not recommend to use QuickProp without modifications in contemporary architectures and tasks.

\bibliographystyle{plain}
\bibliography{document}

\end{document}